\documentclass[11pt,letterpaper,logo]{mystyle}

\usepackage[numbers]{natbib}
\usepackage{algorithm}
\usepackage{algorithmic}
\usepackage{fontawesome}
\usepackage{fontawesome}

\hypersetup{
  colorlinks=true,
  citecolor=EvolventAccentDark,
  linkcolor=EvolventAccent,
  urlcolor=EvolventAccent
}

\title{G\"odel Forest: Balancing Search Depth and Breadth for Data-Centric Recursive Self-Improvement}
\runningtitle{G\"odel Forest}

\newcommand{\correspondingemailone}{mkhu@evolvent.co}
\newcommand{\correspondingemailtwo}{xiao-ming.wu@polyu.edu.hk}

\author{Ziqi Zhao$^{1,2}$, Fanqing Meng$^{1,3,*}$, Haocheng Lu$^{1,4}$, Lingxiao Du$^{1,3}$, Qiguang Chen$^{1}$,\authorcr
Mengkang Hu$^{1,\dagger}$, Xiao-ming Wu$^{2,\dagger}$\\
{\small $^{1}$Evolvent AI \quad $^{2}$The Hong Kong Polytechnic University} \\ 
{\small$^{3}$National University of Singapore \quad $^{4}$Columbia University }\\
{\small $^{*}$Project leader. $^{\dagger}$Corresponding authors.}\\
{\fontsize{9pt}{12pt}\selectfont
\href{https://github.com/evolvent-ai/Godel-Forest}{\faGithub~\,https://github.com/evolvent-ai/Godel-Forest}\hspace{14pt}%
\href{http://rsibench.co}{\faGlobe~\,rsibench.co}}}

\begin{document}

\begin{abstract}
Recursive self-improvement (RSI) aims to achieve compounding gains by having models improve themselves.
While most existing RSI systems optimize external agent harnesses or prompts around a frozen base model, \textit{data-centric RSI} directly updates the model's own parameters by training on agent-generated data.
However, because validating data strategies requires expensive model training, existing methods face a fundamental dilemma: a single agent gets trapped in narrow directions and lacks exploration breadth, while naive parallel search or heavy trace sharing sacrifices long-horizon search depth.
To address this challenge, we introduce \textbf{G\"odel Forest}, a multi-agent framework that organizes recursive self-improvement as an ensemble of co-evolving search trees.
In G\"odel Forest, each agent autonomously grows a persistent tree, deepening, branching, or pruning data strategies based on model feedback to secure depth, while parallel trees explore distinct regions of the data space to expand breadth.
Crucially, rather than leaving trees isolated or flooding them with heavy execution logs, a dynamically co-evolving memory connects the forest: agents continuously distill their successes and failures into compact procedural lessons anchored to a global leaderboard.
Through this forest ecosystem, a dead-end in one tree instantly warns the whole forest against unpromising paths, while an empirical breakthrough quickly seeds new exploration branches in neighboring trees.
Evaluated on RSIBench-Data across six diverse domains, G\"odel Forest outperforms the single-agent baseline by an average of 10.70\% while reducing wall-clock time on five tasks.
Ablations confirm that co-evolving shared memory yields a +7.00\% gain over independent parallel search, demonstrating that collective distillation is key to scalable self-improvement.
The code is available at \url{https://github.com/evolvent-ai/Godel-Forest}.
\end{abstract}

\maketitle
\thispagestyle{firststyle}
\begingroup
\renewcommand{\thefootnote}{\ensuremath{\dagger}}
\footnotetext{Corresponding authors: \href{mailto:\correspondingemailone}{\correspondingemailone};
\href{mailto:\correspondingemailtwo}{\correspondingemailtwo}.}
\endgroup

\section{Introduction}

Recursive self-improvement (RSI) refers to a process in which a system repeatedly produces improved versions of itself under a given evaluator,
thereby improving its own ability to improve and yielding compounding gains~\citep{li2026towards,li2026ft,yu2026autotrainess,zhao2026andes}.
The original G\"odel machine casts RSI as a proof-guided search over self-modifications,
accepting a candidate rewrite only after proving that it increases expected utility.
Because this proof search is intractable for realistic agents,
recent empirical methods~\citep{robeyns2025self,zhang2026darwin,wang2026huxley} replace formal proof with
validation based on observed downstream task utility.
Recent extensions such as the Mendel G\"odel Machine (MGM)\citep{liu2026mendel} further enrich this loop by contrasting execution traces across tasks and parallel runs to evolve executable agent harnesses.

\begin{figure}[!t]
\centering
\includegraphics[width=0.9\linewidth]{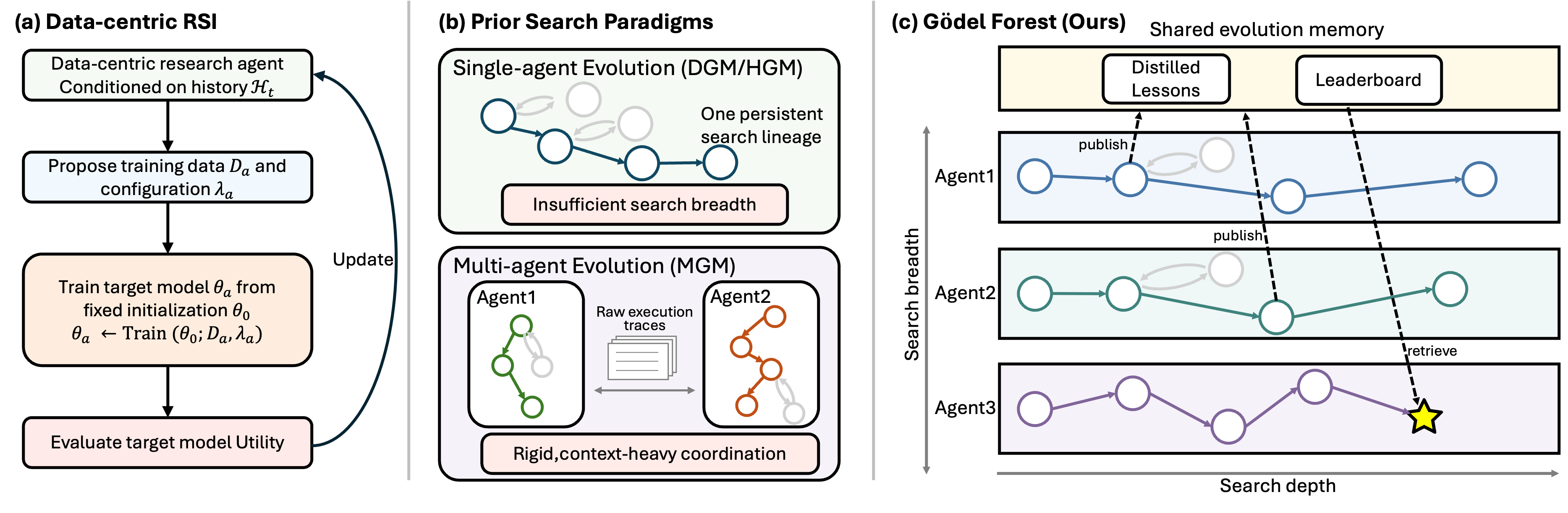}
\vspace{-6pt}
\caption{Comparison of search paradigms in recursive self-improvement (RSI).
}
\label{fig:intro-comparison}
\vspace{-10pt}
\end{figure}

However, RSI systems remain confined to the external harness or prompts around a frozen base model, rather than entering the model itself through parameter updates.
Data-centric RSI offers a complementary route in which an LLM agent repeatedly
proposes data and training strategies, trains candidate target models from fixed
initial parameters, and evaluates them on downstream tasks
\citep{meng2026rsibench,kulikov2026autodata,rank2026posttrainbench,yano2025lamdagent,kessler2025towards}.
The resulting evidence updates the agent's persistent search state and guides
subsequent proposals.
Thus, training changes the parameters of each candidate target model, whereas
recursion lies in the agent's evidence-driven evolution of data and training
strategies~(Figure~\ref{fig:intro-comparison}a).

Unlike agent harness optimization, data evolution navigates an explosive combinatorial space spanning individual sample synthesis and dataset-level distribution shifts.
Crucially, validating each candidate hypothesis incurs full-loop model training and downstream evaluation, rendering naive exploration prohibitively expensive under constrained compute budgets.
This training-in-the-loop bottleneck traps existing RSI paradigms in a stark dilemma~(Figure~\ref{fig:intro-comparison}b):
single-agent frameworks~\citep{zhang2026darwin,wang2026huxley} easily stagnate in myopic, narrow trajectories lacking directional breadth, whereas multi-agent variants~\citep{liu2026mendel} rely on raw trace exchange across identical tasks, resulting in rigid, context-saturated coordination that fails to scale to open-ended data discovery.
How to sustain deep, long-horizon hypothesis pursuit while unlocking expansive exploration breadth through lightweight coordination remains the central challenge of data-centric RSI.

To resolve this, we propose \textbf{G\"odel Forest}~(Figure~\ref{fig:intro-comparison}c), a multi-agent framework that structures RSI as an ensemble of co-evolving search trees.
G\"odel Forest reconciles the depth-versus-breadth trade-off through a structured duality:
locally, each agent autonomously cultivates a deep, persistent search tree, using target-model feedback to continuously deepen, refine, or backtrack its data strategies to secure long-horizon search depth;
globally, parallel trees explore distinct regions of the combinatorial data space, opening expansive directional breadth without discarding their accumulated search context.
Crucially, rather than leaving trees isolated or choking communication with raw execution traces, agents coordinate asynchronously via a dynamically co-evolving shared memory through a read-evolve-publish protocol.
By distilling empirical breakthroughs and failure modes into compact, actionable data lessons anchored to a global leaderboard, G\"odel Forest establishes a collective feedback loop:
a dead-end hit by one tree instantly acts as an inductive guardrail across the forest, while an empirical leap swiftly seeds new exploration branches in neighboring trees.

We evaluate G\"odel Forest on RSIBench-Data across six diverse benchmarks spanning repository-level software engineering (SWE-bench Verified, Multilingual, and Pro), long-horizon terminal interaction (Terminal-Bench~2.0), scientific reasoning (GPQA Diamond), and competition mathematics (AIME 2026).
Across all tasks, G\"odel Forest consistently outperforms the single-agent baseline, achieving an average gain of 10.70\% (e.g., boosting SWE-bench Verified from 35.00\% to 54.00\%) while reducing end-to-end wall-clock time on five of the six benchmarks.
Ablations further confirm that co-evolving via shared evolution memory provides a +7.00\% gain over running independent parallel agents, demonstrating that dynamic experience sharing is key to scalable self-improvement.

The contributions of this paper are as follows:
\begin{itemize}[leftmargin=*,topsep=0pt]
\item We formulate data-centric RSI as an open-ended data-evolution search, identifying the fundamental tension between long-horizon search depth and directional exploration breadth under expensive training-in-the-loop validation.
\item We propose \textbf{G\"odel Forest}, a co-evolutionary multi-agent framework that structures recursive improvement as an ensemble of persistent search trees, coupling private long-horizon tree refinement with an asynchronous read-evolve-publish shared memory for lightweight coordination.
\item We evaluate \textbf{G\"odel Forest} across six diverse benchmarks in RSIBench-Data, demonstrating consistent gains over single-agent and independent parallel baselines, superior wall-clock efficiency, and significant empirical advantages from shared-memory co-evolution.
\end{itemize}

\section{Related Work}
\label{sec:related}

\paragraph{Empirical G\"odel Machines and Harness Evolution.}
Originating from Schmidhuber's theoretical vision~\citep{schmidhuber2006goedelmachinesselfreferentialuniversal}, empirical G\"odel machines replace intractable formal proof searches with observed task utility to iteratively rewrite agent programs~\citep{robeyns2025self,yin2025godel}.
Subsequent efforts generalize this paradigm through archive-based exploration~\citep{zhang2026darwin,wang2026huxley}, risk-controlled acceptance~\citep{wu2025sgm}, evaluator co-evolution~\citep{zhang2026hyperagents,iacob2026red}, and multi-trajectory comparative evolution~\citep{liu2026mendel,qu2026coralautonomousmultiagentevolution}.
Despite these advances, existing empirical G\"odel systems operate under a shared constraint: their self-modifications are strictly restricted to external executable harnesses or prompt wrappers around frozen base models, precluding internal parameter updates.
Furthermore, exchanging verbose execution traces induces severe context bloat, preventing effective scaling in open-ended research spaces.

\paragraph{Data-Centric Recursive Self-Improvement.}
A parallel line of research pursues compounding gains by directly updating model parameters.
Early frameworks bootstrap reasoning capabilities from self-generated rationales~\citep{zelikman2022star,chen2024self,yuan2024self} or zero-data environment self-play~\citep{zhao2026absolute,huang2026r}.
Building on interactive environments~\citep{khan2025dataenvgym,kessler2025towards}, recent post-training frameworks delegate dataset curation, recipe synthesis, and execution monitoring to autonomous LLM agents~\citep{yano2025lamdagent,li2026ft,meng2026rsibench}.
However, existing data-centric RSI systems predominantly rely on single-agent search that easily gets trapped in localized data distributions under constrained budgets, while uncoordinated parallel sampling repeatedly wastes compute on redundant failure modes.

Unlike prior work confined to prompt-level harnesses or uncoordinated sampling, we establish \textit{asynchronous memory co-evolution} as the coordination paradigm for data-centric RSI.
G\"odel Forest realizes this through a search duality: private search trees maintain local selective pressure for deep hypothesis refinement, while a shared memory substrate prevents redundant exploration and propagates breakthrough recipes across workers.
Grounded in real parameter updates, G\"odel Forest ultimately bridges empirical G\"odel search with scalable post-training collective intelligence.

\section{G\"odel Forest}
\label{sec:method}

To overcome the training-in-the-loop bottleneck in data-centric recursive self-improvement (RSI), we present \textbf{G\"odel Forest}, a co-evolutionary multi-agent framework that structures search as an ensemble of persistent search trees coupled through a dynamically shared evolution memory.
G\"odel Forest resolves the fundamental tension between search depth and exploration breadth through a structured duality: locally, individual agents cultivate deep, persistent search trees with budget-aware diagnostic gating; globally, parallel trees explore distinct regions of the combinatorial data space and coordinate asynchronously via compact, distilled empirical lessons.

\subsection{Problem Formulation and Single-Tree Foundations}
\label{sec:method-formulation}

\paragraph{Empirical G\"odel search as tree search.}
The original G\"odel machine~\citep{schmidhuber2006goedelmachinesselfreferentialuniversal} executes self-modifications only after formally proving that they yield higher expected utility than continuing the search.
Because formal proof search is intractable for complex generative agents, empirical G\"odel frameworks guide exploration using observed downstream utility over an archive of agent versions.
Let $\mathcal{T}_t$ denote the agent archive at step $t$, initialized with a seed agent as $\mathcal{T}_0=\{a_0\}$.
Each non-root agent retains the identity of the parent from which it was derived, forming a directed tree.
A search policy selects an archived agent $a \in \mathcal{T}_t$ and applies a modification operator $a' \leftarrow \Phi(a, \mathcal{H}_t)$ conditioned on the accumulated evaluation evidence $\mathcal{H}_t$.
The successor $a'$ is appended to the archive as a child of $a$, giving $\mathcal{T}_{t+1} = \mathcal{T}_t \cup \{a'\}$.
Because any archived node can be selected for modification, the search can either deepen a promising branch or backtrack to an earlier node to pursue an alternative direction.

\paragraph{Data-centric recursive self-improvement.}
Unlike prior empirical G\"odel systems that optimize external prompt templates or agent harnesses around frozen base models, data-centric RSI seeks compounding gains by directly updating target model parameters.
Here, each archived agent $a$ is a specialized data-engineering program.
Executing $a$ synthesizes training data $D_a$, trains candidate parameters $\theta_a$ from a fixed base model $\theta_0$, and evaluates downstream utility $U(a)$ on dataset $\mathcal{B}$:
\begin{align}
  (D_a, \lambda_a) &= \operatorname{Generate}(a), \label{eq:agent-generation} \\
  \theta_a &= \operatorname{Train}(\theta_0; D_a, \lambda_a), \label{eq:target-training} \\
  U(a) &= \operatorname{Eval}_{\mathcal{B}}(\theta_a). \label{eq:prelim-utility}
\end{align}
Although every candidate model is trained from the same initial parameters $\theta_0$, the process remains strictly recursive: the empirical performance and failure diagnostics of earlier target models update the agent's persistent state, which in turn governs subsequent data generation and curation decisions.

\paragraph{The depth-versus-breadth dilemma.}
Validating each data hypothesis requires full-loop model training and downstream evaluation, rendering naive exploration prohibitively expensive.
This training-in-the-loop bottleneck creates a stark dilemma: a single tree easily stagnates in path-dependent trajectories lacking directional breadth, whereas parallel search either redundantly repeats shared failures or chokes on verbose trace exchange~\citep{liu2026mendel}.
Reconciling deep, multi-round hypothesis pursuit with broad, lightweight collective coordination is the central objective of G\"odel Forest.

\subsection{Persistent Tree Search: Local Deep Hypothesis Pursuit}
\label{sec:method-local}

To enable long-horizon hypothesis pursuit without premature abandonment, each worker $i \in \{1, \dots, K\}$ maintains a private search tree $\mathcal{T}_t^{(i)}$ and evidence ledger $\mathcal{H}_t^{(i)}$.
Each worker LLM autonomously selects parents, designs data and training configurations, and makes other search decisions under fixed interfaces and budgets.
Within each tree, the worker executes a skill-enhanced cognitive loop comprising four distinct stages:

\paragraph{Stage 1: Diagnose.}
The worker analyzes the failure modes of previous attempts in $\mathcal{H}_t^{(i)}$ (e.g., runaway completion, ungrounded actions, or skipped verification) to localize specific behavioral defects in the candidate model.

\paragraph{Stage 2: Propose.}
Conditioned on both its private trajectory $\mathcal{H}_t^{(i)}$ and the globally shared memory $M_t$, the worker selects an existing parent agent $a \in \mathcal{T}_t^{(i)}$ and generates an evolved successor program $a'$ (realized as executable data synthesis and curation code):
\begin{equation}
  a' \leftarrow \Phi(a, \mathcal{H}_t^{(i)}, M_t).
  \label{eq:dual-conditioned-phi}
\end{equation}
The successor $a'$ is archived in $\mathcal{T}_t^{(i)}$, executing $(D_{a'}, \lambda_{a'}) = \operatorname{Generate}(a')$ followed by candidate model training $\theta_{a'} = \operatorname{Train}(\theta_0; D_{a'}, \lambda_{a'})$.

\paragraph{Stage 3: Validate \& Evaluate.}
Full-benchmark evaluation on long-horizon tasks is computationally demanding.
To preserve the evaluation budget, the worker first validates the generated data and evaluates the candidate on a lightweight diagnostic subset $\mathcal{B}^{\mathrm{diag}} \subset \mathcal{B}$, whose composition is determined autonomously by the worker:
\begin{equation}
  U_{\mathrm{diag}}(a') = \operatorname{Eval}_{\mathcal{B}^{\mathrm{diag}}}(\theta_{a'}).
  \label{eq:diagnostic-utility}
\end{equation}
If $U_{\mathrm{diag}}(a')$ surpasses the gating threshold, the worker proceeds to evaluate the remaining tasks in $\mathcal{B}$ using the same checkpoint $\theta_{a'}$ without retraining, obtaining the complete utility $U(a')$.
Otherwise, evaluation is terminated early, pruning unviable candidates before incurring full-set overhead while recording the diagnostic evidence.

\paragraph{Stage 4: Review.}
The worker appends the evaluation outcome to $\mathcal{H}_t^{(i)}$ and determines the next search action: exploiting the current direction through further modification, revising hyperparameters, or backtracking to an earlier ancestor node.

\begin{algorithm}[!t]
\caption{G\"odel Forest Asynchronous Co-Evolution Protocol}
\label{alg:godel-forest}
\small
\begin{algorithmic}[1]
\STATE \textbf{Initialize:} Archives $\mathcal{T}_0^{(i)} \leftarrow \{a_0\}$, evidence $\mathcal{H}_0^{(i)} \leftarrow \emptyset$ for $i \in \{1,\dots,K\}$, shared memory $M_0 \leftarrow (\emptyset, \emptyset)$.
\WHILE{elapsed wall-clock $W \le W_{\max}$ \textbf{and} cumulative expenditure $C \le C_{\max}$}
  \FOR{each worker $i \in \{1, \dots, K\}$ \textbf{in parallel, asynchronously}}
    \STATE $\mathcal{L}_{\mathrm{rel}}^{(i)} \leftarrow \operatorname{Retrieve}(\mathcal{L}_t, \mathcal{H}_t^{(i)})$ \COMMENT{\textbf{Read}: Query relevant shared lessons}
    \STATE Select parent $a \in \mathcal{T}_t^{(i)}$ and propose $a' \leftarrow \Phi(a, \mathcal{H}_t^{(i)}, \mathcal{L}_{\mathrm{rel}}^{(i)}, \Lambda_t)$ \COMMENT{\textbf{Propose}}
    \STATE $(D_{a'}, \lambda_{a'}) \leftarrow \operatorname{Generate}(a')$; \quad $\theta_{a'} \leftarrow \operatorname{Train}(\theta_0; D_{a'}, \lambda_{a'})$ \COMMENT{\textbf{Train}}
    \STATE $U_{\mathrm{diag}}(a') \leftarrow \operatorname{Eval}_{\mathcal{B}^{\mathrm{diag}}}(\theta_{a'})$ \COMMENT{\textbf{Diagnostic Screening}}
    \IF{$U_{\mathrm{diag}}(a')$ passes gating threshold}
      \STATE $U_{\mathrm{rem}}(a') \leftarrow \operatorname{Eval}_{\mathcal{B}\setminus\mathcal{B}^{\mathrm{diag}}}(\theta_{a'})$; \quad $U(a') \leftarrow \operatorname{Merge}(U_{\mathrm{diag}}(a'), U_{\mathrm{rem}}(a'))$
      \STATE $\Lambda_t \leftarrow \Lambda_t \cup \{(a', U(a'))\}$ \COMMENT{Full eval \& leaderboard update}
    \ENDIF
    \STATE Update private evidence $\mathcal{H}_{t+1}^{(i)}$; \quad Distill lesson $\ell_{a'}^{(i)} \leftarrow \operatorname{Synthesize}(\mathcal{H}_{t+1}^{(i)})$ \COMMENT{\textbf{Review}}
    \IF{$\ell_{a'}^{(i)}$ contains non-trivial empirical findings}
      \STATE $\mathcal{L}_{t+1} \leftarrow \mathcal{L}_t \cup \{\ell_{a'}^{(i)}\}$ \COMMENT{\textbf{Publish}: Broadcast lesson to blackboard}
    \ENDIF
  \ENDFOR
\ENDWHILE
\STATE \textbf{return} Optimal agent $a^\star = \arg\max_{a \in \Lambda_{t_{\mathrm{end}}}} U(a)$ with target checkpoint $\theta_{a^\star}$.
\end{algorithmic}
\end{algorithm}

\subsection{Forest Co-Evolution via Shared Evolution Memory}
\label{sec:method-memory}

To reconcile independent exploration with collective synergy, G\"odel Forest organizes $K$ concurrent workers into a co-evolving ensemble $\mathcal{F}_t = \{\mathcal{T}_t^{(i)}\}_{i=1}^K$.
Workers interact asynchronously through a shared evolution memory $M_t = (\Lambda_t, \mathcal{L}_t)$ that acts as a non-blocking public blackboard.

\paragraph{Outcome-level anchor: shared leaderboard $\Lambda_t$.}
The leaderboard aggregates verified utilities across all fully evaluated agents across workers, $\Lambda_t = \{ (a, U(a)) \mid a \in \bigcup_{j=1}^{K}\mathcal{T}_t^{(j)},\ a\ \text{fully evaluated} \}$, establishing a transparent global performance anchor for each worker to gauge whether its local branch remains competitive against the collective frontier.

\paragraph{Procedural cognition: structured empirical lessons $\mathcal{L}_t$.}
While numeric scores indicate performance, cross-tree coordination requires understanding \emph{why} a data intervention succeeded or failed.
Instead of broadcasting raw data samples or execution traces, worker $i$ synthesizes each significant trial into a compact, structured lesson $\ell$:
\begin{equation}
  \ell = \big\langle \mathcal{H}_{\text{intent}},\, \Delta U,\, \text{Attribution},\, \rho \big\rangle,
  \label{eq:lesson-schema}
\end{equation}
where $\mathcal{H}_{\text{intent}}$ specifies the tested data hypothesis (e.g., all-turn vs.\ terminal-turn supervision), $\Delta U$ records the empirical utility shift, $\text{Attribution}$ denotes the causal diagnosis of the outcome, and $\rho$ articulates a distilled, actionable operational rule (e.g., ``restrict supervision loss to final assistant turns to prevent run-on completion'').
This structured schema decouples high-level procedural takeaways from heavy execution logs, keeping cross-tree communication lightweight ($\sim$100 tokens per lesson) and circumventing the severe context bloat that impedes raw-trace sharing~\citep{liu2026mendel}.
The published lessons dynamically form an asynchronous shared blackboard: $\mathcal{L}_{t+1} = \mathcal{L}_t \cup \{ \ell_a^{(i)} \}$.

\paragraph{Asynchronous Read-Evolve-Publish loop.}
Communication in G\"odel Forest is entirely pull-based and non-blocking, operating without global synchronization barriers:
\begin{itemize}[leftmargin=*,topsep=0pt,itemsep=0pt,parsep=0pt]
  \item \textbf{Read}: Prior to formulating a new candidate, worker $i$ retrieves contextually relevant lessons based on its current failure symptoms:
  $\mathcal{L}_{\mathrm{rel}}^{(i)} = \operatorname{Retrieve}(\mathcal{L}_t, \mathcal{H}_t^{(i)}) \subseteq \mathcal{L}_t$.
  \item \textbf{Evolve}: The retrieved lessons fulfill two complementary functions: negative lessons serve as \emph{inductive guardrails}, pruning paths that caused failure in peer trees; positive lessons act as \emph{catalytic seeds}, inspiring peer workers to explore creative variants in distinct domains.
  \item \textbf{Publish}: Upon concluding its Review stage, the worker broadcasts any distilled lesson to $\mathcal{L}_t$, immediately enriching the collective memory for all subsequent queries.
\end{itemize}
Algorithm~\ref{alg:godel-forest} summarizes the complete G\"odel Forest co-evolutionary protocol.
Upon reaching budget termination, the manager returns the globally optimal checkpoint $\theta_{a^\star}$ satisfying $a^\star = \arg\max_{a \in \Lambda} U(a)$.
This architecture ensures that individual trees maintain deep, uninterrupted search continuity, while shared memory continuously drives collective discovery across the forest.

\section{Experiments}
\label{sec:experiments}

\subsection{Experimental Setup}
\label{sec:experimental-setup}

\paragraph{Benchmarks and task interface.}
We evaluate G\"odel Forest on RSIBench-Data~\citep{meng2026rsibench} across six diverse, high-difficulty benchmarks: SWE-bench Verified~\citep{jimenez2024swe}, SWE-bench Multilingual~\citep{zan2026multi}, SWE-bench Pro~\citep{deng2025swe}, Terminal-Bench~2.0~\citep{merrill2026terminal}, GPQA Diamond~\citep{rein2023gpqa}, and AIME~2026.
Together, these cover repository-level software engineering, long-horizon interactive bash environments, multidisciplinary scientific reasoning, and competition mathematics.
The three SWE-bench suites use the Mini-SWE-Agent runner~\citep{yang2024swe}.
All candidates are evaluated by Harbor in isolated E2B sandboxes under identical interfaces.

\paragraph{Models, training infrastructure, and budgets.}
Every candidate starts from the base model \texttt{Qwen3.5-35B-\allowbreak A3B-Base}~\citep{yang2025qwen3} and is fine-tuned with LoRA SFT through the shared \textsc{Tinker} backend; Claude Opus~4.8 serves as the fixed external rollout generator.
Each run is strictly governed by a 16-hour wall-clock limit and a \$500 cumulative \textsc{Tinker} budget, enforced across the entire forest.
\paragraph{Official evaluation and baselines.}
During search, $U(a)$ denotes feedback from the RSIBench-Data selection evaluator, which guides data generation and checkpoint selection without exposing gold solutions. Post-search, the selected checkpoint $\theta_{a^\star}$ is frozen and officially evaluated in an isolated canonical run, yielding $S_{\mathrm{off}} = \operatorname{Eval}^{\mathrm{off}}(\theta_{a^\star})$. Because both phases share task identities, $S_{\mathrm{off}}$ serves as a standardized post-selection verification rather than a held-out generalization benchmark.
Our primary baseline is vanilla Claude Code\footnote{\url{https://github.com/anthropics/claude-code}} with \texttt{Sonnet-5} operating as a single-agent search without forest coordination or shared memory.
To demonstrate framework generality across underlying LLM backbones, we additionally instantiate G\"odel Forest with OpenAI Codex\footnote{\url{https://github.com/openai/codex}} (\texttt{gpt-5.6-sol}).
Claude Code runs at high reasoning effort and Codex at maximum reasoning effort, following RSIBench-Data standards.

\subsection{Main Results}
\label{sec:main-results}

Table~\ref{tab:main-results} presents the official benchmark performance, end-to-end wall-clock time, and cumulative \textsc{Tinker} expenditure for G\"odel Forest ($K=3$) versus the vanilla Claude Code baseline.

\begin{table}[!t]
\caption{Official performance and resource use with Claude Code + \texttt{Sonnet-5}. Time is end-to-end wall-clock hours for data generation, training, validation, and review; cost is cumulative \textsc{Tinker} expenditure across all workers. Here the official scores $S_{\mathrm{off}}$ are reported. }
\vspace{2pt}
\label{tab:main-results}
\centering
\begin{tabular}{@{}llrrr@{}}
\toprule
Benchmark & Method & Score $\uparrow$ & Time (h) $\downarrow$ & Cost (\$) $\downarrow$ \\
\midrule
SWE-bench Verified
  & Vanilla & 35.00\% & 14.91 & \textbf{181.70} \\
  & G\"odel Forest & \textbf{54.00\%} & \textbf{14.06} & 239.67 \\
\addlinespace[1pt]
SWE-bench Multilingual
  & Vanilla & 22.00\% & 14.20 & 363.77 \\
  & G\"odel Forest & \textbf{30.00\%} & \textbf{11.02} & \textbf{336.55} \\
\addlinespace[1pt]
SWE-bench Pro
  & Vanilla & 4.00\% & 9.54 & \textbf{45.63} \\
  & G\"odel Forest & \textbf{15.00\%} & \textbf{7.49} & 419.87 \\
\addlinespace[1pt]
Terminal-Bench~2.0
  & Vanilla & 5.62\% & \textbf{8.87} & \textbf{156.93} \\
  & G\"odel Forest & \textbf{17.98\%} & 15.50 & 295.06 \\
\addlinespace[1pt]
GPQA Diamond
  & Vanilla & 52.00\% & 6.43 & 16.03 \\
  & G\"odel Forest & \textbf{55.00\%} & \textbf{5.76} & \textbf{15.22} \\
\addlinespace[1pt]
AIME 2026
  & Vanilla & 49.17\% & 9.29 & 121.21 \\
  & G\"odel Forest & \textbf{60.00\%} & \textbf{8.61} & \textbf{49.94} \\
\bottomrule
\end{tabular}
\end{table}

\paragraph{Universal gains across diverse scenarios.}
G\"odel Forest consistently outperforms the single-agent baseline on all six benchmarks, yielding an average absolute gain of 10.70\%.
The most pronounced improvement emerges on SWE-bench Verified, soaring from 35.00\% to 54.00\% (+19.00\%).
Substantial gains also extend across challenging executable domains (SWE-bench Pro improves from 4.00\% to 15.00\%; Terminal-Bench~2.0 improves from 5.62\% to 17.98\%) and complex reasoning tasks (GPQA Diamond reaches 55.00\%; AIME 2026 reaches 60.00\%).
These comprehensive improvements confirm that co-evolving persistent search trees consistently unlocks superior data synthesis strategies across heterogeneous problem distributions.

\paragraph{The efficiency paradox: faster completion via asynchronous synergy.}
Despite conducting broader and deeper multi-tree exploration, G\"odel Forest achieves \emph{lower} end-to-end wall-clock time on five of the six benchmarks (e.g., reducing SWE-bench Multilingual from 14.20h to 11.02h, and SWE-bench Pro from 9.54h to 7.49h).
This efficiency stems directly from two algorithmic properties: asynchronous parallel execution maximizes hardware concurrency, while shared evolution memory rapidly propagates negative findings, enabling workers to prune unviable exploration branches before committing expensive training cycles.

\paragraph{Understanding inference expenditure.}
While wall-clock time decreases, cumulative expenditure is higher on long-horizon agentic benchmarks (e.g., SWE-bench Pro and Terminal-Bench~2.0).
This expenditure disparity is an artifact of candidate capability rather than search inefficiency: weaker baseline models frequently crash, loop, or terminate prematurely during early turns, truncating their interaction trajectories.
In contrast, models evolved by G\"odel Forest sustain coherent, multi-turn reasoning and tool use, executing complex actions that incur higher inference cost but successfully solve challenging tasks.


\subsection{Cross-Backbone Generalization: Results with Codex}
\label{sec:codex-results}

\begin{figure}[!t]
\centering
\includegraphics[width=0.88\linewidth]{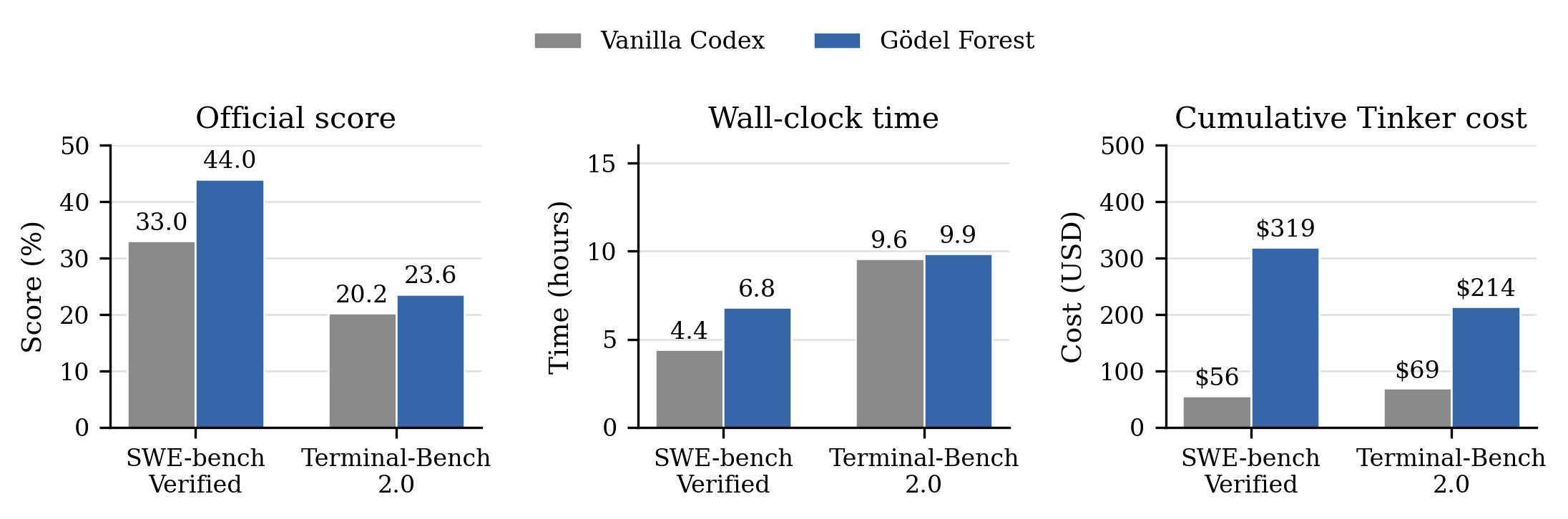}
\caption{Cross-backbone generalization with Codex (\texttt{gpt-5.6-sol}). Left: official score. Center: wall-clock time. Right: cumulative \textsc{Tinker} expenditure.}
\label{fig:model-comparison}

\end{figure}

To confirm that G\"odel Forest functions as a general meta-architecture independent of any specific agent implementation, we evaluate worker agents driven by Codex (\texttt{gpt-5.6-sol}) on SWE-bench Verified and Terminal-Bench~2.0 (Figure~\ref{fig:model-comparison}).
With target model, training backend, and evaluation interfaces held fixed, G\"odel Forest secures marked improvements over the vanilla Codex baseline: official scores rise from 33.00\% to 44.00\% (+11.00\%) on SWE-bench Verified and from 20.22\% to 23.60\% (+3.38\%) on Terminal-Bench~2.0.
Furthermore, elapsed wall-clock latency remains modest (6.81h vs.\ 4.41h on SWE-bench Verified, and 9.85h vs.\ 9.56h on Terminal-Bench~2.0).
As in Section~\ref{sec:main-results}, the higher expenditure (\$319.22 vs.\ \$55.61) reflects the target model's enhanced ability to sustain extended, successful problem-solving interactions.
These consistent gains demonstrate that the benefits of asynchronous tree co-evolution generalize across distinct frontier LLMs.

\subsection{Ablation Study: Dissecting Exploration Breadth and Coordination}
\label{sec:ablation-study}

\begin{figure}[!t]
\centering
\includegraphics[width=0.88\linewidth]{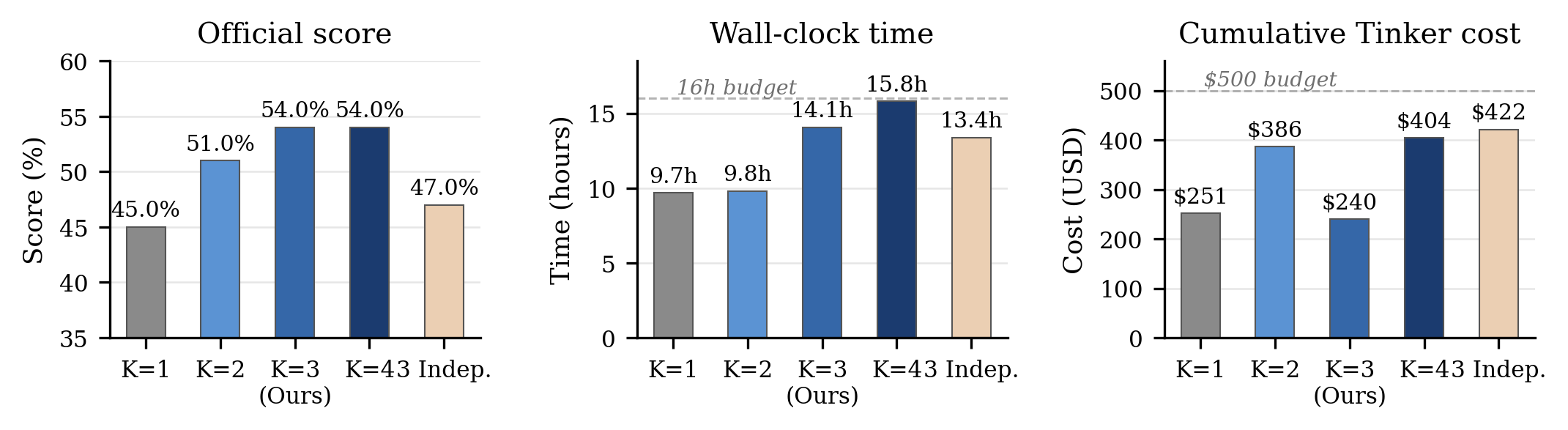}
\caption{Ablation study on SWE-bench Verified. We evaluate worker scaling ($K \in \{1, 2, 3, 4\}$) with full shared memory, and compare against 3 Independent Workers without shared memory. Dashed lines denote 16-hour and \$500 budget limits.}
\vspace{-10pt}
\label{fig:ablation}

\end{figure}

We conduct an ablation study on SWE-bench Verified to isolate the contributions of tree concurrency and shared memory coordination (Figure~\ref{fig:ablation}).\vspace{-6pt}

\paragraph{Concurrency scaling and the depth-breadth trade-off.}
Scaling the number of persistent workers from $K=1$ to $K=3$ broadens initial exploration diversity, steadily elevating official performance from $45.00\%$ to $51.00\%$ and $\mathbf{54.00\%}$.
Here, the single-worker setting ($K=1$) isolates our local persistent tree search (\S\ref{sec:method-local}) from forest coordination.
Unlike the vanilla baseline operating over an unconstrained flat conversational loop, a worker in our framework incorporates a structured evidence ledger $\mathcal{H}$ for systematic backtracking, cognitive stage skills for grounded candidate proposal, and diagnostic gating ($\mathcal{B}^{\mathrm{diag}} \to \mathcal{B}$) to prune unviable recipes early.
However, scaling further to $K=4$ plateaus at $54.00\%$ while substantially increasing wall-clock time (15.80h) and cost (\$404.43).
Under a fixed total budget, excessive worker concurrency splinters compute across too many concurrent branches, diminishing the search depth achievable along any individual tree.
Thus, $K=3$ establishes an optimal Pareto balance between exploration breadth, refinement depth, and resource expenditure.\vspace{-6pt}

\paragraph{Collective synergy vs.\ uncoordinated concurrency.}
Contrasting G\"odel Forest with the \textbf{3~Indep.}\ baseline (three concurrent workers evolving without shared memory) isolates the effect of cross-tree communication.
Deprived of shared lessons and the global leaderboard, three independent workers achieve only 47.00\% (a steep 7.00\% performance drop) despite consuming comparable budget (\$421.63 and 13.38h).
Without shared memory, parallel workers repeatedly squander training resources on identical dead-ends that were already diagnosed by peers.
This confirms that G\"odel Forest's superior performance stems from collective cognitive synergy rather than naive parallel compute scaling.

\begin{figure}[!t]
\centering
\includegraphics[width=0.82\linewidth]{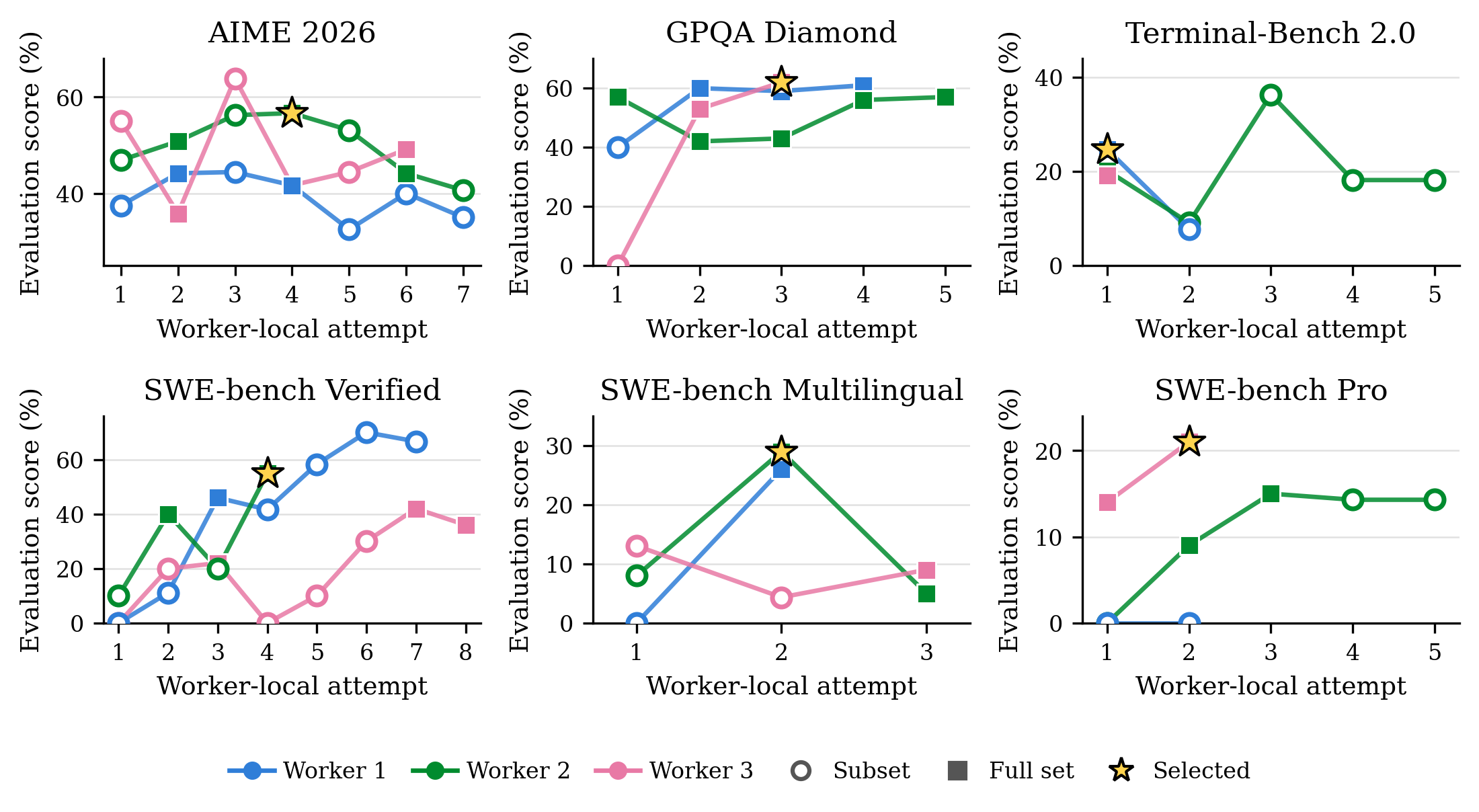}
\vspace{-10pt}
\caption{Candidate search trajectories across six benchmarks. Colors trace independent trees maintained by the three workers ($K=3$). Open circles denote diagnostic evaluations ($\mathcal{B}^{\mathrm{diag}}$); filled squares denote full benchmark evaluations ($\mathcal{B}$). Yellow stars mark the selected checkpoints.}
\vspace{-10pt}

\label{fig:attempt-trajectories}

\end{figure}

\subsection{Search Dynamics and Cross-Worker Knowledge Synthesis}
\label{sec:search-dynamics-synthesis}

\begin{figure}[!t]
\centering
\includegraphics[width=0.76\linewidth]{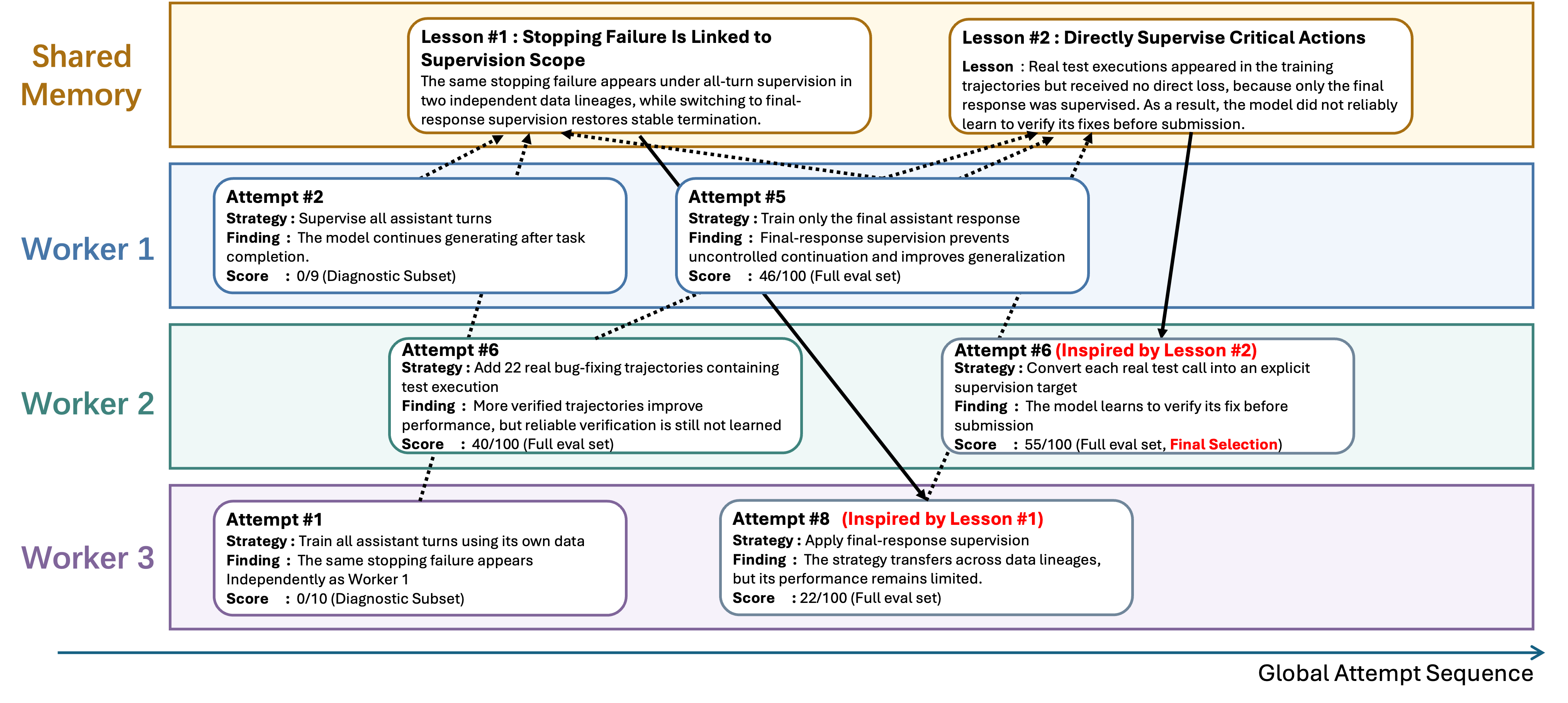}
\vspace{-10pt}
\caption{Cross-worker knowledge synthesis on SWE-bench Verified. Workers cultivate private search trees (horizontal lanes) and coordinate asynchronously via shared evolution memory (top lane). Dotted arrows trace lesson distillation from empirical findings; dashed arrows show lessons inspiring subsequent data interventions. Candidate scores reflect within-run evaluations.}
\vspace{-10pt}
\label{fig:case-study}

\end{figure}

To understand how persistent tree search and shared memory interact in practice, we examine both the macroscopic trajectory dynamics across all benchmarks (Figure~\ref{fig:attempt-trajectories}) and a microscopic case study on SWE-bench Verified (Figure~\ref{fig:case-study}).\vspace{-6pt}

\paragraph{Macroscopic search dynamics: volatility and multi-tree resilience.}
As illustrated in Figure~\ref{fig:attempt-trajectories}, data-centric RSI trajectories are inherently non-monotonic: high-performing checkpoints are frequently followed by sharp performance drops when workers explore aggressive data mixtures or novel supervision formats.
Rather than guaranteeing monotonic ascent, each attempt acts as an empirical hypothesis test.
In this volatile optimization landscape, G\"odel Forest provides critical structural resilience:
if an individual worker hits a severe regression or local dead-end, parallel workers continue advancing complementary branches; meanwhile, the persistent global archive decouples speculative exploration from checkpoint selection, guaranteeing that temporary failures never forfeit prior performance peaks.\vspace{-6pt}

\paragraph{Microscopic knowledge synthesis: turning failures into collaborative leaps.}
Figure~\ref{fig:case-study} illustrates how workers coordinate without copying raw data or sharing verbose logs, but by synthesizing empirical breakdowns into compact, actionable lessons.\vspace{-6pt}

\paragraph{Lesson~\#1 (Stopping Boundary Guardrail):}
Under dense all-turn supervision, Worker~1 (Attempt~\#2, $0/9$) and Worker~3 (Attempt~\#1, $0/10$) independently encounter severe runaway generation where the model fails to terminate after finishing tasks.
Diagnosing that multi-turn loss degrades termination tokens, Worker~1 restricts supervision strictly to the final assistant turn in Attempt~\#5, restoring termination boundaries and achieving a $46\%$ score.
This finding crystallizes into \textbf{Lesson~\#1}: \emph{loss masking to the final assistant turn restores stable task termination}.
Reading this lesson, Worker~3 immediately adopts final-turn masking in Attempt~\#8, boosting its score from near-zero to $22\%$ and preventing further compute waste.\vspace{-6pt}

\paragraph{Lesson~\#2 (Supervising Verification Actions):}
Although final-turn supervision stabilizes completion, it exposes an unexpected limitation: the model ceases executing self-verification tests because intermediate bash calls receive zero training loss.
Worker~2 notes this in Attempt~\#6 ($40\%$), distilling \textbf{Lesson~\#2}: \emph{critical intermediate actions, such as test executions, require explicit supervision}.
To reconcile this with Lesson~\#1, Worker~2 synthesizes \emph{prefix-truncated trajectories}, converting intermediate verification commands into pseudo-terminal targets.
In revised Attempt~\#6, the model actively executes test suites prior to submission, achieving the peak score of $\mathbf{55\%}$ and earning final checkpoint selection.

Together, these empirical dynamics show how G\"odel Forest turns noisy trial-and-error into compounding, self-correcting scientific discovery.

\section{Conclusion}

We presented \textbf{G\"odel Forest}, a multi-agent framework for data-centric recursive self-improvement that resolves the depth-versus-breadth search dilemma.
Locally, agents cultivate persistent search trees with diagnostic gating for deep hypothesis pursuit; globally, parallel trees explore distinct data regions and coordinate asynchronously via shared memory.
By distilling outcomes into compact shared lessons anchored to a leaderboard, G\"odel Forest establishes a collective feedback loop where failures prune dead-ends and breakthroughs seed new branches.
Across six RSIBench-Data benchmarks, G\"odel Forest achieves consistent gains over single-agent and parallel baselines (+10.70\% average score, +7.00\% ablation gain) with lower wall-clock latency, validating asynchronous tree co-evolution as a scalable self-improvement paradigm.



\bibliographystyle{unsrtnat}
\bibliography{ref}

\end{document}